\documentclass[11pt]{article}

\usepackage{deauthor}
\usepackage{times}
\usepackage{authblk}
\usepackage{graphicx}

\usepackage{float,color}
\usepackage{makecell}
\usepackage{comment}
\usepackage{algorithmic}
\usepackage{algorithm}
\usepackage{amsfonts}  
\usepackage{array}
\usepackage{xurl}
\usepackage{subfig}
\usepackage{caption}
\usepackage{enumitem,kantlipsum}
\usepackage{placeins}
\usepackage{amsmath}
\usepackage{amssymb}
\usepackage{scalerel}
\usepackage{dsfont}
\usepackage{mathtools}
\usepackage{bbm}
\usepackage{diagbox}
\usepackage{chngpage}
\usepackage{subfig}
\usepackage{booktabs}
\usepackage{xcolor}
\usepackage{bbding}
\usepackage{pifont}

\DeclareMathOperator*{\argmin}{argmin} 
\DeclareMathOperator*{\argmax}{argmax} 
\newcommand{\xhdr}[1]{{\vspace{1pt}\noindent\bfseries #1}.}
\newcommand{\ie}{\textit{i.e., }}
\newcommand{\eg}{\textit{e.g., }}

\newcommand{\etc}{\textit{etc.}}
\newcommand{\wrt}{\textit{w.r.t. }}

\newcommand{\aka}{\textit{aka. }}

\definecolor{citecol}{HTML}{2DDC0E}
\definecolor{tableofcontent}{HTML}{E63E15}
\definecolor{urlcol}{HTML}{2470D8}
\usepackage{hyperref}
\hypersetup{
    colorlinks=true,       % false: boxed links; true: colored links
    linkcolor=tableofcontent, 
    citecolor=citecol,        % color of links to bibliography
    urlcolor=black,           % color of external links
}

\begin{document}

\title{Generative Explanations for Graph Neural Network: \\Methods and Evaluations}

%%%%%
\author{Jialin Chen$^\dagger$ ,\quad Kenza Amara$^{\ddagger}$,\quad Junchi Yu$^\S$ ,\quad Rex Ying$^\dagger$ \\
    $\dagger$ Department of Computer Science, Yale University\\
     \texttt{\small \{jialin.chen, rex.ying\}@yale.edu}\\
    $\ddagger$ Department of Computer Science, ETH Zurich\\
     \texttt{\small kenza.amara@ai.ethz.ch}\\
    $\S$ Institute of Automation, Chinese Academy of Sciences\\
    \texttt{\small yujunchi2019@ia.ac.cn}\\
    }

\maketitle
\begin{abstract}
       Graph Neural Networks (GNNs) achieve state-of-the-art performance in various graph-related tasks. However, the black-box nature often limits their interpretability and trustworthiness. Numerous explainability methods have been proposed to uncover the decision-making logic of GNNs, by generating underlying explanatory substructures. In this paper, we conduct a comprehensive review of the existing explanation methods for GNNs from the perspective of graph generation. Specifically, we propose a unified optimization objective for generative explanation methods, comprising two sub-objectives: Attribution and Information constraints. We further demonstrate their specific manifestations in various generative model architectures and different explanation scenarios. With the unified objective of the explanation problem, we reveal the shared characteristics and distinctions among current methods, laying the foundation for future methodological advancements. Empirical results demonstrate the advantages and limitations of different explainability approaches in terms of explanation performance, efficiency, and generalizability.
\end{abstract}
\section{Introduction}
Graph Neural Networks (GNNs) have emerged as a powerful tool for studying graph-structured data in various applications, including social networks, drug discovery, and recommendation systems~\cite{GNN_application, GNN_anomaly, GNN_financial, GNN_financial_time_series,GNN_molecule, GNN_molecule_review, GNN_socialnetwork}. The explainability and trustworthiness of GNNs are crucial for their successful deployment in real-world scenarios, especially in high-stake applications, such as anti-money laundering, fraud detection, and healthcare forecasting~\cite{Graph-SST2,explainability_GCN, Evaluate_explainability}. Explanations for GNNs aim to discover the reasoning logic behind their predictions, making them more understandable and transparent to users. Explanations also help identify potential biases and build trust in the decision-making process of the model. Furthermore, they aid users in understanding complex graph-structured data and improve outcomes in various applications through better feature selection~\cite{survey_zhang,Dai:survey,Survey_trustworthy}.

Numerous explanation methods have been extensively studied for GNNs, including gradient-based attribution methods~\cite{Grad-CAM-Graph, SA-Graph, IG}, perturbation-based methods~\cite{GNNExplainer, ReFine, SubgraphX,GraphMask, GraphLIME}, \textit{etc.} However, most of these methods optimize individual explanations for a specific instance, lacking global attention to the overall dataset and the ability to generalize to unseen instances. To tackle this challenge, generative explainability methods have emerged recently, which instead formulate the explanation task as a distribution learning problem. Generative explainability methods aim to learn the underlying distributions of the explanatory graphs across the entire graph dataset~\cite{RCExplainer, GFlowExplainer, CLEAR, XGNN}, providing a more holistic approach to GNN explanations.

Current surveys in the field of Graph Neural Networks (GNNs) explainability primarily focus on the taxonomy and evaluation of explanation methods~\cite{Graph-SST2, Evaluate_explainability, Survey_counterfactual}, as well as broader trustworthy aspects such as robustness, privacy, and fairness~\cite{survey_cyber, survey_zhang, Survey_li, Survey_trustworthy}. 
%However, these surveys lack a comprehensive and systematic overview of the strengths and limitations associated with these explainability methods. In particular, 
The emerging generative explainability methods prompt us to consider the potential advantages of incorporating distribution learning into the optimization objective, such as better explanation efficiency and generalizability.

% \rex{the however statement is a bit too broad. we can mention more concrete advantages of a generative approach that these surveys didn't investigate, such as generalization}

Our work stands apart from previous works by thoroughly investigating the different mechanisms for generating explanations. We explore a comprehensive range of graph generation techniques that have been employed in GNN explanation tasks, including cutting-edge techniques such as the Variational Graph Autoencoder (VGAE) and denoising diffusion models. Our study elucidates the core design considerations of different generative models and unifies a group of effective GNN explanation approaches from a novel generative perspective. The key insight lies in a unified optimization objective, including an \textit{Attribution} constraint and an \textit{Information} constraint to ensure that the generated explanations are sufficiently succinct and relevant to the predictions. We then delve into the details of the practical designs of the \textit{Attribution} and \textit{Information} constraints to facilitate the connections and potential extensions of current generative explainability methods. Moreover, the proposed framework guides GNN users to efficiently design effective generative explainability methods for practical use.

Comprehensive experiments on synthetic and real-world datasets demonstrate the advantages and drawbacks of these existing methods. Specifically, our results show that generative approaches are empirically more efficient during the inference stage. Meanwhile, generative approaches achieve the best generalization capacity compared to other non-generative approaches.

This paper is structured as follows. First, we introduce the notations and problem setting in Sec.~\ref{sec:pre}. Then, we propose a standard optimization objective with \textit{Attribution} constraint and \textit{Information} constraint to unify generative explanation methods in Sec.~\ref{sec:generative_formulation}. Detailed expressions of these two constraints are elaborated in Sec.~\ref{sec:generative_model} and Sec.~\ref{sec:info_constraint}, respectively. In Sec.~\ref{sec:extention_case}, we discuss how to generalize the proposed framework to extensive explanation scenarios, \eg counterfactual and model-level explanations. Additionally, we present a taxonomy of representative works with different generative backbones in Sec.~\ref{sec:taxonomy}. Finally, we conduct comprehensive evaluations and demonstrate the potential of deep generative methods for GNN explanation in Section~\ref{sec:experiment}.

% \rex{A paragraph summarizing the results}

% GNNs are widely used. Explanations for GNNs are essential in applications (cite). Common usage of GNN explainability focuses on instance-level, model-level, and counterfactual explanations.

% Many explainability methods have been developed (cite). However, they utilize a variety of approaches, such as ... (also cite some papers that are not generative models), and thelacksack a unified framework to describe the most effective approaches for GNN explanations under the instance-level, model-level and counterfactual settings. (use a figure to illustrate the challenge)

% There have been extensive survey studies regarding GNN explainability methods. CITE categorizes the methods into XXX. ... (how we are different) They do not investigate the desired properties of high quality GNN explanations such as XXX

% Here we unify a group of effective GNN explainability approaches through a novel generative perspective. The key insight is (sth about Eq 1 / 2)...
% The objective contains XX components: ...

% Advantages of our framework (how we addressed the challenges of many explainability methods). Key insights we can extract from comparison of various generative models for generating explanations.
% \rex{can we claim that we are the focusing on deep explanation models? Are there explainability methods using deep learning but does not use a generative approach?}

% Evaluation.

\section{Preliminaries}\label{sec:pre}
% In this section, we first define the notations used for GNN explanation tasks and introduce the definitions of explanation graphs in different scenarios
% in Sec.~\ref{sec:notation}. Then we introduce the problem setting of explanation tasks in Sec.~\ref{sec:general_formulation} and  from a generative perspective in Sec.~\ref{sec:generative_formulation}.
\subsection{Notations and Definitions} \label{sec:notation}
Given a well-trained GNN model $f$ (\aka base model) and an instance (\ie a node or a graph) of the dataset, the objective of the explanation task is to identify concise graph substructures that contribute the most to the model's predictions. The given graph (or $N$-hop neighboring subgraph of the given node) can be represented as a quadruplet $G(\mathcal{V},\mathcal{E}, \mathbf{X}, \mathbf{E})$, where $\mathcal{V}$ is the node set, $\mathcal{E}\subseteq \mathcal{V}\times \mathcal{V}$ is the edge set. $\mathbf{X}\in\mathbb{R}^{|\mathcal{V}|\times d_n}$ and $\mathbf{V}\in\mathbb{R}^{|\mathcal{E}|\times d_e}$ denote the feature matrices for nodes and edges, respectively, where $d_n$ and $d_e$ are the dimensions of node features and edge features. In this work, we focus on structural explanation, \ie we keep the dimensions of node and edge features unchanged. The main notations used throughout this paper are summarized in Table~\ref{tab:notation}. Depending on the specific explanation scenario, we define the explanation graphs as follows. 
\begin{table}[th]
    \centering
    \begin{tabular}{c|c}
    \toprule
            \textbf{Notation} & \textbf{Description}\\
            \midrule
         $G(\mathcal{V},\mathcal{E}, \mathbf{X}, \mathbf{E})$& a graph with nodes $\mathcal{V}$, edges $\mathcal{E}$, node features $\mathbf{X}$ and edge features $\mathbf{E}$ \\
         $ \mathcal{Y}$ & the label space\\
         $ f$ & the well-trained GNN to be explained (base model)\\
         $G_e$ & an explanation graph\\
         $G_{ce}$ & a counterfactual explanation graph\\
         $G_{m}^c$& a model-level explanation graph for the target class $c$ \\
         $\mathcal{G}=\{G^1, \cdots, G^M\}$ & the graph set of $M$ input instances\\
         $\mathcal{G}_e=\{G_e^1, \cdots, G_e^M\}$ & the set of $M$ generated explanation graphs\\
         $ Y_f(s)\in\{0,1,\cdots, |\mathcal{Y}|\}$ &the predicted label of the instance $s$ by $f$\\
         $ Y^\ast\in\{0,1,\cdots, |\mathcal{Y}|\}$ & the target predicted label during the explanation stage\\
         $ P_f(s)\in[0,1]^{|\mathcal{Y}|}$ &the output probability vector of the instance $s$ by $f$\\
         $ f(s)\in\mathbb{R}^{|\mathcal{Y}|}$ & the output logit vector of the instance $s$ by $f$\\
         $ g_\theta(\cdot): \mathcal{G}\rightarrow \mathcal{G}_e$ & an explanation generator with parameters $\theta$ \\
         % $ q_\phi(\cdot): \mathcal{G}\rightarrow \mathbf{Z}$ &Encoder network with parameters $\phi$\\
         % $ p_\theta (\cdot): \mathbf{Z}\rightarrow \mathcal{G}_e$ & Decoder network with parameters $\theta$\\
         $ p(\cdot| G)$ &the distribution of the explanation graphs for a given $G$\\
         % $ d(\cdot, \mathcal{G})$ & Function of distance from graph $\mathcal{G}$\\
         % $d_0; d_l; d_1 $ &input; latent; and final embedding dimension of nodes\\
         % $ Z\in\mathbb{R}^{|\mathcal{V}|\times d_l}$ & Node latent embedding matrix\\
         % $ H\in\mathbb{R}^{|\mathcal{V}|\times d_1}$ & Node final embedding matrix\\
         % $v_i\in\mathcal{V}$& Graph node\\
         % $\mathbf{Z}$ & Latent space\\
    \bottomrule
    \end{tabular}
    \caption{Summary of the main notations}
    \label{tab:notation}
\end{table}

\begin{definition}[Explanation Graph]
Given a well-trained GNN $f$ and an instance represented as $G(\mathcal{V},\mathcal{E},\mathbf{X},\mathbf{V})$, an explanation graph $G_{e}(\mathcal{V}_e,\mathcal{E}_e, \mathbf{X_e}, \mathbf{E_e})$ for the instance is a compact subgraph of $G$, such that $\mathcal{V}_e\subseteq \mathcal{V}$, $\mathcal{E}_e\subseteq \mathcal{E}$, $\mathbf{X_e}=\{X_j|v_j\in \mathcal{V}_e\}$ and $\mathbf{E_e}=\{E_k|e_k\in\mathcal{E}_e\}$, where $v_j$ and $X_j$ denote the graph node and the corresponding node feature, $e_k$ and $E_k$ denote the graph edge and the corresponding edge feature. Explanation graph $G_e$ is expected to be compact and result in the same predicted label $Y^\ast$ as the label of $G$ made by $f$, \ie $Y^\ast = Y_f(G_e)=Y_f(G)$, where $Y_f(\cdot)$ denotes the predicted label made by the model $f$.
\end{definition}

\begin{definition}[Counterfactual Explanation Graph]
Given a well-trained GNN $f$ and an instance $G$, a counterfactual explanation graph $G_{ce}$ is as close as possible to the original graph $G$, while it results in a different predicted label $Y^\ast$ from the label of $G$ predicted by $f$, \ie $Y^\ast = Y_f(G_{ce})\neq Y_f(G)$.
\end{definition}

\begin{definition}[Model-level Explanation Graph]
Given a set of graph $\mathcal{G}=\{G^1,\cdots, G^M\}$, where each $G^j\in\mathcal{G}$ has the same label $c$ predicted by the well-trained GNN $f$, a model-level explanation graph $G_m^c$ is a distinctive subgraph pattern that commonly appears in $\mathcal{G}$, and is predicted as the same  label $c$, \ie $Y^\ast=Y_f(G_{m}^c)=c$.
\end{definition}

\subsection{General Explanation for Graph Neural Network}\label{sec:general_formulation}
Given a graph $G$ and the corresponding label $Y^\ast$ in specific explanation scenarios, generating explanation graphs can be formulated as an optimization problem that maximizes the mutual information between the generated graph and the target label $Y^\ast$~\cite{GNNExplainer}, \ie $G^\ast_e =\argmax_{G_e} MI(Y^\ast, G_e) =  \argmin_{G_e}-\mathbb{E}_{Y^{\ast}|G_e}\log P(Y^\ast| G_e)$,
where $MI(\cdot, \cdot)$ denotes the mutual information function. 

\xhdr{Instance-dependent Explainers}
Early efforts develop instance-dependent explainers for GNNs that optimize an explanation for each given instance. For example, the gradient-based explainers~\cite{SA-Graph, Grad-CAM-Graph} evaluate the node and edge importance with the gradient norm of prediction \wrt node and edge features. Other explainers utilize more advanced learning-based approaches such as mask optimization~\cite{GNNExplainer}, surrogate model~\cite{PGM-Explainer}, and Monte Carlo Tree Search~\cite{SubgraphX} to extract the explanation subgraphs for each individual instance.

Although instance-dependent explainers partly reveal the behavior of GNNs, there are several limitations. Firstly, since these methods optimize explanations for individual graphs, they result in significant computation and lower explanation efficiency. Furthermore, the learnable modules within an instance-dependent explainer cannot be applied to explain the predictions for unseen instances, since the parameters are specific for a single instance, leading to worse generalization capacity and a lack of holistic knowledge across the entire dataset.

\section{Generative Framework for Graph Explanations}
\subsection{Unified Optimization Objective}\label{sec:generative_formulation}

To overcome the aforementioned limitations, recent research has developed approaches that leverage deep generative methods to explain GNNs. Instead of optimizing an explanation for individual instances, the generative methods aim to generate explanations for new graphs by learning a strategy to generate the most explanatory subgraphs across the whole dataset. Formally, given a set of input graphs $\mathcal{G}$, the generative explainer learns the distribution of the underlying explanation graphs $p(G_{e}|G)$ using a parameterized subgraph generator $g_{\theta}:\mathcal{G}\rightarrow \mathcal{G}_{e}$. After training, the subgraph generator is capable of identifying the explanation subgraphs that are most important to the target labels:
\begin{equation}\label{eq:objective1}
     \theta^\ast = \argmax_{\theta}\log P_{Y^\ast}(G_e|\theta, G),
\end{equation} 
where $P_{Y^\ast}(G_e|\theta, G)$ is the probability that the generated $G_e=g_\theta(G)$ is a valid explanation for the target label $Y^\ast$.
In addition to the validity requirement, an ideal explanation graph should be sparse and compact compared with the given graph. Directly optimizing Eq.~\ref{eq:objective1} leads to a trivial solution where $G_{e}=G$, as the input graph is most informative for the target label $Y^\ast$.
To obtain a compact explanation, we impose an information constraint $\mathcal{L}_{\textsc{INFO}}(G_e, G)$ that restricts the amount of information contained in the generated explanation subgraph, thereby ensuring the conciseness and brevity of the explanations. The overall optimization objective becomes
\begin{equation}\label{eq:objective2}
    \min_{\theta} -\log P_{Y^\ast}(G_e|\theta, G)+ \mathcal{L}_{\textsc{INFO}}(G_e, G):=\mathcal{L}_{\textsc{ATTR}}(G_e,Y^\ast) + \mathcal{L}_{\textsc{INFO}}(G_e, G).
\end{equation}
We name the first term in Eq.~\ref{eq:objective2} the attribution loss $\mathcal{L}_{\textsc{ATTR}}(G_e, Y^\ast)$, which measures how much $G_e$ captures the most important substructures for the target label $Y^\ast$. $\mathcal{L}_{\textsc{ATTR}}$ is typically the cross-entropy loss for categorical $Y^\ast$ and mean squared loss for continuous $Y^\ast$.

\xhdr{Connection With Variational Auto-encoder} 
The variational constraint can be one of the manifestations of
$\mathcal{L}_{\textsc{INFO}}$ in Eq.~\ref{eq:objective2}, \ie $\mathcal{L}_{\textsc{INFO}}(G, G_e):=\mathrm{D_{KL}}\left(q_\theta(G_e |  G) \| Q(G_e)\right)$, where $\mathrm{D_{KL}}$ denotes Kullback–Leibler divergence, $Q(G_e)$ is the prior distribution of the generated explanation graph $G_e$, and $q_\theta(G_e|G)$ is the variational approximation to $g_\theta(G_e| G)$, variational constraint drives the posterior distribution of $G_e$ generated by $g_\theta(\cdot|G)$ to its prior distribution, thus restricting the information contained in $G_e$ in the process. The overall objective is 
\begin{equation}\label{eq:objective_vc}
    \mathcal{L}=\mathbb{E}_G-\log P_{Y^\ast}(G_e | \theta, G) + \mathrm{D_{KL}}\left(q_\theta(G_e |  G) \| Q(G_e)\right),
\end{equation}
In this case, the objective of generative explanation shares similar spirits with the Variational Auto-encoder (VAE)~\cite{GVAE}.
Recall the optimization objective of VAE is
\begin{equation}\label{eq:vae}
    \mathcal{L}_{\textsc{VAE}}=\mathbb{E}_{z \sim q_\phi(z | G)} -\log \left(p_\varphi(G | z)\right)+\mathrm{D_{KL}}\left(q_\phi(z| G) \| q (z)\right),
\end{equation}
where $q_\phi$ is the encoder that maps graph $G$ into a latent space, then the decoder $p_\varphi$ recovers the original graph $G$ based on the latent representation $z$. $q (z)$ is the prior distribution of the latent representation, which is usually a Gaussian distribution. Notably, the generative explanation approach with the variational constraint as $\mathcal{L}_{\textsc{INFO}}$ (Eq.~\ref{eq:objective_vc}) is a variant of Variational Auto-encoder (VAE) (Eq.~\ref{eq:vae}), albeit with two fundamental distinctions. Firstly, VAE aims to learn the distribution of the original graph, whereas generative explanation focuses on learning the underlying distribution of explanatory structures. Secondly, VAE constrains the distribution of latent representation $z$, while the generative explanation constrains the posterior distribution of $G_e$. These distinctions highlight the methodologies for generalizing generative models to the task of GNN explainability.

\subsection{Taxonomies of Generative Models}\label{sec:generative_model}
In this section, we will discuss several taxonomies of generative models that have been employed in the field of GNN explainability. These models aim to learn the probability distribution of the underlying explanatory substructures by training across the entire graph dataset.
\paragraph{Mask Generation (MG)~\cite{PGExplainer, yugraph,GSAT,gnninterpreter}}
The mask generation method is to optimize a mask generator $g_{\theta}$ to generate the edge mask $M$ for the input graph $G$. The elements of the mask represent the importance score of the corresponding edges, which is further employed to select the important substructures as explanations for the input graph. The mask generator is usually a graph encoder followed by a multi-layer perception (MLP), which first embeds edge representations $h_{e_{i}}$ for each edge and then generates the sampling probability $p_{i}$ for edge $e_{i}$. The mask $m_{i}\in \{0,1\}$ of $e_{i}$ is sampled from the Bernoulli distribution $\mathrm{Bern}(p_{i})$. Since the sampling process is non-differentiable, the Gumbel-Softmax trick is usually employed for continuous relaxation as follows:

\begin{equation}\label{eq:gumbel-softmax}
    m_{i}=\sigma ((\log{\epsilon}-\log{(1-\epsilon)}+\log(p_{i})-\log(1-p_i))/\tau),\quad \epsilon \sim \mathrm{Uniform}(0,1)
\end{equation}
where $\tau$ is the temperature, $\sigma$ is the sigmoid function. When $\tau$ goes to zero, $m_i$ is close to the discrete Bernoulli distribution. The explanation $G_{e}$ is obtained by applying the edge mask $M$ to the input graph $G$, \ie $G_{e}=M\odot G=g_\theta(G)\odot G$, where $\odot$ is element-wise multiplication.
Given an input graph $G$ and the target label $Y^\ast$, the parameter $\theta$ of the mask generator $g_{\theta}$ is optimized by minimizing the following attribution loss:

\begin{equation}\label{eq:MO}
\mathcal{L}_{\textsc{ATTR}} =-\mathbb{E}_{G}\log P_{Y^\ast}(G_e|G,\theta) \hbox{~~with~~} G_e=g_\theta(G)\odot G,
 \end{equation}
which is typically the cross entropy between the output probability $P_f(G_e)$ made by $f$ and the target label $Y^\ast$.

\paragraph{Variational Graph Autoencoder (VGAE)~\cite{GEM,CLEAR,OrphicX}}
Variational Graph Autoencoder (VGAE)~\cite{GVAE} is a variational autoencoder for graph-structured data, where the encoder $q_\phi(\cdot)$ and the decoder $p_\theta(\cdot)$ are typically parameterized by graph neural networks. VGAE can be used to learn the distribution of the underlying explanations and thus generate explanation graphs for unseen instances. The encoder maps an input graph $G$ to a probability distribution over a latent space. The decoder then samples from the latent space and recovers an explanation graph by $G_e=p_{\theta}(z)$. The attribution loss of VGAE for generating explanation graphs is
\begin{equation}\label{eq:loss_GVAE}
    \mathcal{L}_{\textsc{ATTR}}=\mathbb{E}_{z\sim q_\phi(z| G)}-\log P_{Y^\ast}(G_e|\theta, z, G)+\mathrm{D_{KL}}(q_\phi(z | G) \| q(z)).
\end{equation}
The standard VGAE shown in Eq.~\ref{eq:vae} aims to generate realistic graphs. On the contrary, the VGAE-based explainer (Eq.~\ref{eq:loss_GVAE}) maximizes the likelihood of the valid explanation graph $G_e$ for the target label $Y^\ast$. The former term in Eq.~\ref{eq:loss_GVAE} evaluates whether the explanation graph $G_e$ captures the most important structures for $Y^\ast$. CLEAR~\cite{CLEAR} sets the first term as the cross entropy between the output probabilities $P_f(G_e)$ and $Y^\ast$, while GEM~\cite{GEM} utilizes the cross entropy between the generated graph $G_e$ and the ground-truth explanation. The second term of KL divergence is a model-specific constraint that drives the posterior distribution $q_\phi(z  | G)$ to the prior distribution $q(z )$, which is usually a Gaussian distribution. 
\paragraph{Generative Adversarial Networks (GAN)~\cite{Gan-Explainer}}
Generative Adversarial Network (GAN) is a type of generative model that does not include an explicit encoder
component. Instead, GANs consist of a generator $g_\theta$ that creates an explanation graph $G_e=g_\theta(z)$ for $z$ sampled from a prior distribution $q(z)$, and a discriminator $d_\phi$ that distinguishes between the input graph $G$ and the explanation graph $G_e$. The objective function of a GAN is a min-max game in which the generator $g_\theta$ tries to minimize while the discriminator $d_\phi$ tries to maximize it.
\begin{equation}\label{eq:loss_GAN}
    \mathcal{L}_{\textsc{ATTR}} = -\mathbb{E}_{z\sim q(z)}\log P_{Y^\ast}(G_e|\theta, z, G) + \log d_\phi(G) + \mathbb{E}_{z\sim q(z)}\log(1-d_\phi(g_\theta(z))),
\end{equation}
where the first term can be the cross entropy between $P_f(g_\theta(z))$ and the target label $Y^\ast$. $d_\theta(\cdot)$ denotes the probability that the discriminator predicts that the input is an explanation.  GAN-based Explainer~\cite{Gan-Explainer} was recently proposed with the first term replaced with the square of the difference between the output logit of $f$ for $G$ and $G_e$, \ie $\mathbb{E}_{z\sim q(z)} \left(f(G)-f(g_\theta(z))\right)^2$. Once the GAN is well-trained, the generator $g_\theta$ can be employed to generate valid explanation graphs for any unseen instances, given a point in the prior distribution $q(z)$.

\paragraph{Diffusion~\cite{jeanneret2022diffusion, diffusion_visual_explanation, D4Explainer}}The diffusion model is a generative model that has been widely used in graph generation tasks to generate realistic graphs~\cite{vignac2022digress, haefeli2022diffusion}, which contains two key components: the forward diffusion process, and the reverse denoising network. Given an original graph $G_0$, the forward diffusion process progressively adds noise to the initial graph and generates a sequence of noisy graphs $\{G_0, G_1,\cdots, G_T\}$ with increasing levels of noise through $q(G_t|G_0)$. $G_T$ becomes pure noise when $T$ goes to infinity. The reverse denoising network $g_\theta$ learns to remove the noise and recover the target explanation graph by $G_e=g_\theta(G_t)$ for each $t$. Since an ideal $G_e$ is a compact subgraph of the original graph $G$, it is equivalent to ensuring that the complementary subgraph $G-G_e$ is close to $G$. Therefore, we make $G-G_e$ approximate $G$ and take $G_e$ as the output explanation graph for $G$. The loss function is as follows,
\begin{equation}\label{eq:loss_diffusion}
\mathcal{L} = -\mathbb{E}_{t\in[0,T]}\mathbb{E}_{G_t\sim q(G_t|G_0)}\log P_{Y^\ast}(G_e|\theta, G_t, G_0) + \mathbb{E}_{t\in[0,T]}\mathbb{E}_{G_t\sim q(G_t|G_0)}\mathcal{L}_{\textsc{CE}}(G_0-g_\theta(G_t), G_0).
\end{equation}
The second term denotes the binary cross entropy loss across all elements in the adjacency matrices of $(G_0-g_\theta(G_t)$ and $G_0$. Notably, the diffusion-based explainer naturally involves the \textit{Information} constraint into the optimization objective, as $\mathcal{L}_{\textsc{CE}}$ plays the role of $\mathcal{L}_{\textsc{INFO}}$ 
 that restricts the size of the generated explanation graph $G_e$.

\paragraph{Reinforcement Learning Approaches}
Reinforcement Learning (RL) approaches formulate the process of generating an explanation graph as a trajectory of step-wise states. Let $\tau=(s_0, \cdots_, s_K)\in\mathcal{T}$ denote a trajectory $\tau$ that consists of states $s_0, \cdots, s_K$. $\mathcal{T}$ is a set of all possible trajectories. At the $k$-th step, the state $s_k$ refers to a subgraph of the initial graph, denoted as $G_k$. $G_0$ is a starting node from the given graph and $G_K$ is the terminal explanation graph. Let $a_k$ denote the taken action from $s_{k-1}$ to $s_k$, which is usually adding a neighboring edge to the current subgraph $G_{k-1}$. Instead of recovering the distribution of the holistic explanation graphs, reinforcement learning approaches learn a generative agent (policy network) that determines the next action by $a_k\sim g_\theta(G_{k-1})$ with parameters $\theta$. The reward function is a crucial component of reinforcement learning to address the non-differentiability issue of the sampling process within the generative agent. In the explanation task, the reward function measures the benefit of taking the action $a_k$ for the target label $Y^\ast$ with the current subgraph $G_{k-1}$. RCExplainer~\cite{RCExplainer} proposes to take the individual causal effect (ICE)~\cite{glymour2016causal} of the action $a_k$ as the reward. GFlowExplainer~\cite{GFlowExplainer} involves the output probability $P_f(G_k)$ over the target label $Y^\ast$ into the reward design. Reinforcement learning is used in conjunction with another probabilistic model to represent the distribution over the states, \eg Markov Decision Process (MDP), Direct Acyclic Graph (DAG).
\begin{itemize}[leftmargin=*]
\item \xhdr{Markov Decision Process (MDP)~\cite{XGNN, RCExplainer}} The trajectories of states can be framed as a Markov Decision Process. The generative agent $g_\theta(G_{k-1})$ captures the sequential effect of each edge in the generating process toward a target explanation graph. The attribution loss function is as follows,
    \begin{equation}\label{eq:MDP_loss}
        \mathcal{L}_{\textsc{ATTR}}=-\mathbb{E}_{k}[R(G_{k-1}, a_k)]\log P(a_k| \theta, G_{k-1}),
    \end{equation} 
    where $R(G_{k-1},a_k)$ is the reward for the action $a_k$ at state $G_{k-1}$ and $P(a_k| \theta, G_{k-1})$ is the probability of yielding $a_k$ from the distribution $g_{\theta}(G_{k-1})$. This loss function encourages the generative agent to attach higher probabilities to the edges that bring larger rewards, thus leading to an ideal explanation.
    \item \xhdr{Direct Acyclic Graph (DAG)~\cite{GFlowExplainer}} GflowNet~\cite{GFlowNet} frames the trajectories of the states as a direct acyclic graph and aims to train a generative policy network where the distribution over the states is proportional to a pre-defined reward function. A concept of \textit{flow} is introduced to measure the probability flow along the trajectories. It satisfies that the inflows of a state $s_k$ equals the outflows of $s_k$. The attribution loss is as follows,
    \begin{equation}\label{eq:DAG_loss}
        \mathcal{L}_{\textsc{ATTR}}(\tau)=\sum_{s_{k+1}\in\tau}\left(\sum_{(s_k,a_k)\rightarrow s_{k+1}}F(s_k, a_k)-\mathbbm{1}_{s_{k+1}=s_K}R(s_K) - \mathbbm{1}_{s_{k+1}\neq s_K}\sum_{a_{k+1}}F(s_{k+1},a_{k+1})\right)^2,
    \end{equation}
    where $\sum_{(s_k,a_k)\rightarrow s_{k+1}}F(s_k, a_k)$ and $\sum_{a_{k+1}}F(s_{k+1},a_{k+1})$ denote the inflows and outflows of a state $s_{k+1}$, respectively. $\mathbbm{1}$ is used to check whether $s_{k+1}$ is the terminal state $s_K$.  $R(s_K)$ is the reward of the graph $G_K$ corresponding to the terminal state $s_K$. It is provable that the probability distribution of the terminal states generated by the agent trained with Eq.~\ref{eq:DAG_loss} is proportional to their rewards.
\end{itemize}
Typically, reinforcement learning approaches do not rely on an explicit $\mathcal{L}_{\textsc{INFO}}$ to constrain the volume of the explanation graph. Instead, we stop the process of sampling $a_k\sim g_{\theta}(G_{k-1})$ once the stopping criteria are attained, \eg $k$ achieves the pre-defined size of explanation graphs. 

% create trajectories $\tau=(s_0, \cdots, s_K)$ by iteratively sampling $a_k\sim g_{\theta}(G_{k-1})$ and stop this process once the stopping criteria are attained, \eg $K$ achieves the pre-defined size of explanation graphs. 

%\subsection{Taxonomies of Mutual Information Estimations}
% We discuss three categories of subgraph information constraint.
\subsection{Taxonomies of Information Constraint}\label{sec:info_constraint}

Only maximizing the likelihood of the explanatory subgraph with Eq.~\ref{eq:objective1} typically leads to a trivial solution of the whole input graph, which is unsatisfactory. An ideal explanatory subgraph is supposed to have a small portion of the original graph information as well as be faithful for the prediction.
Hence, existing methods~\cite{GSAT,yugraph} introduce an additional information constraint $\mathcal{L}_{\textsc{INFO}}$ as a regularization term to restrict the information of the generated explanation apart from the attribution loss $\mathcal{L}_{\textsc{ATTR}}$. 
The information constraint $\mathcal{L}_{\textsc{INFO}}$ can be categorized as Size Constraint, Mutual Information Constraint, and Variational Constraint.

\paragraph{Size Constraint \cite{GNNExplainer}}
The size constraint is a straightforward approach to restricting subgraph information. Given an input graph $G$ and the size tolerance $K\in (0,|G|)$, the size constraint is 
 $|G_{e}|\leq K$. Here, $|\cdot|$ denotes the volume of a graph, and $K$ is an integer hyperparameter to constrain the volume of explanatory subgraphs. Since applying the same size constraint to different graph sizes is problematic, 
some work employs the sparsity constraint $k\in (0,1)$ and constrains the volume of explanatory subgraph with $|G_{e}|\leq k\cdot |G|$. Recent works~\cite{CLEAR, OrphicX} further utilize a soft size constraint, \ie $\mathcal{L}_{\textsc{INFO}}=d(G, G_e)$, where $d(G, G_e)$ is the distance (\eg L1 and L2 distance) between the adjacency matrices of $G$ and $G_e$.
 
Although the size constraint is intuitive, it has several limitations. Firstly, the topological size is insufficient in measuring the subgraph information as they ignore the information within node and edge features. Secondly, one has to choose different sparsity tolerance $k$ to achieve the best explanation performance, which is difficult due to the trade-off between the sparsity and validity of the explanations. 
% \jialin{discuss $d(G, G-G_e)$ and $d(G,G_{ce})$, which should be a soft size constraints?}

\paragraph{Mutual Information Constraint \cite{yugraph}}
The mutual information constraint restricts the subgraph information by reducing the relevance between the original graphs and the explanatory subgraphs.
Given the graph $G$ and the explanatory subgraph $G_{e}$, the mutual information constraint is formulated as: \begin{equation}\label{eq:MI-constrain}
\mathcal{L}_{\textsc{INFO}}=MI(G,G_{e})=\mathbb{E}_{p(G)}\mathbb{E}_{p(G_{e}|G)}\log{\frac{p(G_{e}|G)}{p(G_{e})}}.
\end{equation}
Here $MI(x,y)$ is the mutual information of two random variables. Minimizing Eq.~\ref{eq:MI-constrain} reduces the relevance between $G$ and $G_{e}$. Thus, the explanation generator tends to leverage limited input graph information to generate the explanation subgraph. 
Compared with the size constraints, the mutual information constraint is more fundamental in information measurement and flexible to different graph sizes.
However, mutual information is intractable to compute, making the constraint impractical to use. One solution is resorting to computationally expensive estimation techniques, such as the Donsker-Varadhan representation of mutual information~\cite{Mine, yugraph}.

\paragraph{Variational Constraint~\cite{GSAT}} The variational constraint is proposed by deriving a tractable variational upper bound of the mutual information constraint. One can plug a prior distribution $q(G_{e})$ into $MI(G,G_{e})$ as the variational approximation to $p(G_e|G)$:
\begin{equation}\label{eq:variational-constraint}
\begin{aligned}
MI(G,G_{e})&=\mathbb{E}_{p(G)}\mathbb{E}_{p(G_{e}|G)}\log{\frac{p(G_{e}|G)}{p(G_{e})}}=\mathbb{E}_{p(G)}\mathbb{E}_{p(G_{e}|G)}\log{\frac{p(G_{e}|G)}{q(G_{e})}} -\mathrm{D_{KL}}(q(G_{e})\|p(G_{e})) \\
&\leq \mathbb{E}_{p(G)}\mathbb{E}_{p(G_{e}|G)}\log{\frac{p(G_{e}|G)}{q(G_{e})}}
:=\mathcal{L}_{\textsc{VC}}.
\end{aligned}
\end{equation}
Here, the inequality is due to the non-negative nature of the Kullback–Leibler (KL) divergence. The posterior distribution $p(G_{e}|G)=\prod_{i=1}^{N}p(e_{i}|\theta)$ is factorized into the multiplication of the marginal distributions of edge sampling $p(e_{i}|\theta)$, which is parameterized with the generative explanation network $\theta$. The prior distribution $q(G_{e})=\prod_{i=1}^{N}q(e_{i})$ is factorized into the prior distributions of edge sampling $q(e_{i})$. $N$ is the total edge number. In practice, $q(e_{i})$ is usually chosen as the Bernoulli distribution or Gaussian distribution. Thus, the variational constraint is an upper bound of the mutual information $MI(G, G_e)$ that can be simplified as $\mathcal{L}_{\textsc{INFO}}=\mathcal{L}_{\textsc{VC}}= \sum_{i=1}^{N}\mathrm{D_{KL}}(p(e_{i}|\theta)\|q(e_{i}))$.
% \begin{equation}\label{eq:simplified-variational-constraint}
% \begin{aligned}
% \mathcal{L}_{\textsc{INFO}}=\mathcal{L}_{\textsc{VC}}= \sum_{i=1}^{N}\mathrm{D_{KL}}(p(e_{i}|\theta)\|q(e_{i})).
% \end{aligned}
% \end{equation}

\subsection{Extension of Explanation Scenarios}\label{sec:extention_case}

\paragraph{Counterfactual Explanation} Most explanation methods focus on discovering the prediction-relevant subgraph to explain GNNs based on Eq.~\ref{eq:objective1}. 
Although these methods can highlight the important substructures for the predictions, they cannot answer the \textit{counterfactual} problem such as: "Will the removal of certain substructure lead to prediction change of GNNs?"
Counterfactual explanations provide insightful information on how the model prediction would change if some event had occurred differently, which is crucial in some real-world scenarios, \eg drug design and molecular generation~\cite{drug_cf, drug_XAI, molecule_cf}. 
Given an input graph $G$, the goal of the generative counterfactual explanation is to train a subgraph generator $g_{\theta}$ to generate a minimal substructure of the input.
If the substructure is removed, the prediction of GNNs will change the most. 
Formally, the objective for counterfactual explanations is $\mathcal{L}_{\textsc{CF}}=-\log{P_{Y^\ast}(G_{ce}|\theta,G)} + \mathcal{L}_{\textsc{INFO}}(G, \overline{G_{ce}})$.
% \begin{equation}\label{eq:counterfactual}
%     \mathcal{L}_{\textsc{CF}}=-\log{P_{Y^\ast}(G_{ce}|\theta,G)} + \mathcal{L}_{\textsc{INFO}}(G, \overline{G_{ce}})
% \end{equation}
Here, $\overline{G_{ce}}=G-G_{ce}$ denotes the generated substructure, which is also the modification applied to $G$ to obtain a counterfactual explanation. $\mathcal{L}_{\textsc{INFO}}$ constrains the information amount contained in $\overline{G_{ce}}$ to be minimal compared with the input graph $G$.

\xhdr{Connection to Graph Adversarial Attack} The counterfactual explanation methods can capture the vulnerability of GNN's prediction since the counterfactual explanation subgraph leads to the prediction change. This problem setting is similar to graph adversarial attacks as they both aim to alter the prediction behavior of the pre-trained GNN by modifying testing graphs. Recall that graph adversarial attacks modify the node features or graph structures to decrease the average performance of a pre-trained GNN. However, the explanation methods change the prediction of each testing sample by instance-level subgraph deletion instead of decreasing the overall testing performance after a one-time graph modification~\cite{model_attack_cf}.

\xhdr{Connection to Graph Out-of-distribution Generalization}
Deep learning models have been found to rely on spurious patterns in the input to make predictions. These patterns are often unstable under distribution shifts, leading to a drop in performance when testing on out-of-distribution (OOD) data. To address this issue, counterfactual augmentation has been proposed as an effective method for improving the OOD generalization ability of deep learning models. This technique involves minimally modifying the training data to change their labels and training the model with both the original and counterfactually augmented data. For graphs, counterfactually augmented subgraphs can be generated by removing subgraphs to create the complementary subgraph, which is a natural form of counterfactual augmentation. However, this approach has received less attention in the context of graph neural networks, presenting an avenue for future research.

\paragraph{Model-level Explanation}
The goal of model-level explanation in the context of GNNs is to identify important graph patterns that contribute to the decision boundaries of the model. Unlike instance-level explanation, model-level explanation provides insights into the general behavior of the model across a range of input graphs with the same predicted label. A brute-force approach to finding these patterns is to mine the subgraphs that commonly appear in graphs and result in the same predicted label. However, this is computationally expensive due to the exponentially large search space. Recently, generative methods have been proposed to generate model-level explanations, leveraging reinforcement learning~\cite{XGNN}, probabilistic generative models~\cite{gnninterpreter}, \etc. 

In these approaches, a generator function $g_\theta(\cdot)$ is used to generate the model-level explanation $G_m$ for a given predicted label $Y^\ast$ based on a set of graphs $\mathcal{G}$ via $G_m\sim g_\theta(\mathcal{G}, Y^\ast)$. The optimization objective for the generator function is to minimize the negative log-likelihood of the explanation given the graphs, while also ensuring that the generated explanation is a compact and recurrent substructure in the set of input graphs:
\begin{equation}\label{eq:model-level_loss}
    \mathcal{L}=-\log P_{Y^\ast}(G_m|\theta, \mathcal{G}) + \mathcal{L}_{\textsc{INFO}}(G_m, \mathcal{G}).
\end{equation}
The first term in Eq.~\ref{eq:model-level_loss} measures whether $G_m$ captures the most determinant graph patterns for the prediction of $Y^\ast$. The generator $g_\theta(\cdot)$ can be modeled by applicable generative model architectures discussed in Sec.~\ref{sec:generative_model}. $\mathcal{L}_{\textsc{INFO}}(G_m,\mathcal{G})$ ensures that $G_m$ is a compact substructure that commonly appears in $\mathcal{G}$.

\section{Method Taxonomy}\label{sec:taxonomy}
\begin{table*}[htb]
\caption{A comprehensive summary of existing generative explanation methods for Graph Neural Networks. RL-MDP denotes the reinforcement learning approach based on Markov Decision Process and RL-DAG denotes the reinforcement learning approach based on Direct Acyclic Graph.} 
\vspace{-5pt}
\centering
\resizebox{0.95\linewidth}{!}{
\begin{tabular}{lccccc}
\toprule
\textbf{Method} & \textbf{Generator} & \textbf{Information Constraint} & \textbf{Level} & \textbf{Scenario} & \textbf{Output}\\
\midrule
PGExplainer~\cite{PGExplainer}&Mask Generation&size&instance&factual&E\\
GIB~\cite{yugraph}&Mask Generation & mutual information & instance&factual & N\\
GSAT~\cite{GSAT}&Mask Generation&variational&instance&factual&E\\
GNNInterpreter~\cite{gnninterpreter}&Mask Generation&size&model&factual&N / E / NF\\
GEM~\cite{GEM}&VGAE&size&instance&factual&E\\
CLEAR~\cite{CLEAR} & VGAE & size & instance & counterfactual & E / NF\\
OrphicX~\cite{OrphicX} & VGAE & variational \& size  & instance & factual & E\\
D4Explainer~\cite{D4Explainer}&Diffusion&size&instance \& model &counterfactual & E\\
GANExplainer~\cite{Gan-Explainer} & GAN &- & instance&factual&E\\
RCExplainer~\cite{RCExplainer}&RL-MDP&size&instance&factual&\textsc{subgraph}\\
XGNN~\cite{XGNN}&RL-MDP&size&model&factual&\textsc{subgraph}\\
GFlowExplainer~\cite{GFlowExplainer}&RL-DAG&size&instance&factual&\textsc{subgraph}\\
\bottomrule
\end{tabular}}
\label{table:comparison}
\vspace{-0.2cm}
\end{table*}
The information constraints $\mathcal{L}_{\textsc{INFO}}$ in Sec.~\ref{sec:info_constraint} and the attribution constraints $\mathcal{L}_{\textsc{ATTR}}$in Sec.~\ref{sec:generative_model} can be combined to construct an overall optimization objective for GNN explainability. 
We provide a comprehensive comparison and summary of existing generative explanation methods and their corresponding generators and information constraints in Table~\ref{table:comparison}. Most existing approaches focus on instance-level factual explanations, while CLEAR~\cite{CLEAR} focuses on counterfactual explanation and D4Explainer~\cite{D4Explainer} is applicable for both counterfactual and model-level explanations. GIB~\cite{yugraph} proposes to deploy mutual information between the generated explanation graph and the original graph as the information constraint, while GSAT~\cite{GSAT} and OrphicX~\cite{OrphicX} utilize the variational constraint. We further compare the outputs of these approaches (the last column in Table~\ref{table:comparison}), where $\textsc{E}$ denotes outputting edge importance with continuous values, $\textsc{N}$ denotes node importance with continuous values, $\textsc{NF}$ denotes the importance of node features and $\textsc{subgraph}$ denotes hard masks for discrete explanatory subgraphs.
\section{Evaluation}\label{sec:experiment}

\subsection{Experimental setting}
\xhdr{Datasets} We evaluate the explainability methods on both synthetic and real-world datasets in different domains, including MUTAG, BBBP, MNIST, BA-2Motifs and BA-MultiShapes. \textbf{BA-2Motifs}~\cite{PGExplainer} is a synthetic dataset with binary graph labels. The house motif and the cycle motif give class labels and thus are regarded as ground-truth explanations for the two classes. \textbf{BA-MultiShapes}~\cite{bamultishapes} is a more complicated synthetic dataset with multiple motifs. Class 0 indicates that the instance is a plain BA graph or a BA graph with a house, a grid, a wheel, or the three motifs together. On the contrary, Class 1 denotes BA graphs with two of these three motifs. \textbf{MUTAG} is a collection of $\sim$3000 nitroaromatic compounds and it includes binary labels on their mutagenicity on Salmonella typhimurium. The chemical fragments -NO2 and -NH2 in mutagen graphs are labeled as ground-truth explanations~\cite{PGExplainer}. The Blood–brain barrier penetration \textbf{BBBP} dataset includes binary labels for over 2000 compounds on their permeability properties. In molecular datasets, node features encode the atom type and edge features encode the type of bonds that connect atoms. \textbf{MNIST75sp} contains graphs that are converted from images in MNIST~\cite{lecun1998gradient} using superpixels. In these graphs, the nodes represent the superpixels, and the edges are determined by the spatial proximity between the superpixels. The coordinates and intensity of the corresponding superpixel construct the node features. Dataset statistics are summarized in Table~\ref{tab:dataset_performance}.
% We evaluate the explainability methods on molecular datasets, Mutag~\cite{debnath1991structure} and BBBP~\cite{wu2018moleculenet}, and on the synthetic dataset BA-2Motifs~\cite{PGExplainer}. \textbf{Mutag} is a collection of 3000 nitroaromatic compounds and it includes binary labels on their mutagenicity on Salmonella typhimurium. The chemical fragments -NO2 and -NH2 in mutagen graphs are labeled as ground-truth explanations~\cite{PGExplainer}. The Blood–brain barrier penetration \textbf{BBBP} dataset includes binary labels for over 2000 compounds on their permeability properties. In molecular datasets, node features encode the atom type and edge features encode the type of bonds that connect atoms. \textbf{BA-2Motifs} is a synthetic dataset with binary graph labels. The house motif and the cycle motif give class labels and thus are regarded as ground-truth explanations for the two classes.

% All experiments are conducted on a Linux machine with an Nvidia GeForce RTX 2070 SUPER GPU with 8GB memory. CUDA version is 11.1. Methods are implemented with Torch 1.9.1.
\begin{table}[ht]
\centering
\resizebox{0.8\linewidth}{!}{
\begin{tabular}{lccc||cc}
\toprule
                 & \textbf{MUTAG} & \textbf{BBBP} & \textbf{MNIST75sp}  & \textbf{BA-2Motifs} & \textbf{BA-MultiShapes}\\ \hline
\# graphs        & 2,951  & 2,039 & 70,000 & 1,000 &  1,000    \\ 
\# node features & 14    & 9    & 5      & 1    &   10    \\ 
\# edge features & 1     & 3    & 1      & 1    &   1   \\ 
Avg \# nodes     & 30  & 24 & 67     & 25   &    40  \\ 
Avg \# edges     & 61    & 52   & 541    & 51   &     87  \\ 
Avg degree       & 2.0     & 2.1  & 7.9    & 2.0    &    2.2   \\ 
\# classes       & 2     & 2    & 10     & 2    &   2   \\ 
GNN performance  & 0.94      &  0.92    &  0.96      &  1.00    &   0.71    \\
\bottomrule
\end{tabular}}
\caption{Dataset statistics and accuracy performance of the GNN model on the test set}\label{tab:dataset_performance}
\end{table}
\vspace{-0.5cm}
\paragraph{GNN models}  For each dataset, we first train a base GNN. We have tested four GNN models: GCN \cite{GCN}, GIN\cite{GIN}, GAT \cite{GAT}, and GraphTransformer \cite{GraphTransf}. We display the results of GraphTransformer for real-world datasets and GIN for synthetic datasets since they give the highest accuracy scores on the test sets, with a reasonable training time and fast convergence. GraphTransformer and GIN have also the advantage of considering edge features, extending their use to more complex graph datasets. The network structure of the GNN models for graph classification is a series of 3 layers with ReLU activation, followed by a max pooling layer to get graph representations before the final fully connected layer. We split the train/validation/test with 80/10/10$\%$ for all datasets and adopt the Adam optimizer with an initial learning rate of 0.001. Each model is trained for 200 epochs with an early stop. The GNN accuracy performances are shown in Table~\ref{tab:dataset_performance}, which demonstrates that the base GNNs are sufficiently powerful for graph classifications on both synthetic and real-world datasets.

\xhdr{Explainability methods} We compare non-generative methods: Saliency~\cite{SA-Graph}, Integrated Gradient~\cite{IG}, Occlusion~\cite{occlusion}, Grad-CAM~\cite{Grad-CAM-Graph}, GNNExplainer~\cite{GNNExplainer}, PGMExplainer~\cite{PGM-Explainer}, and SubgraphX~\cite{SubgraphX}, with generative ones: PGExplainer~\cite{PGExplainer}, GSAT~\cite{GSAT}, GraphCFE (CLEAR)~\cite{CLEAR}, D4Explainer~\cite{D4Explainer} and RCExplainer~\cite{RCExplainer}. Following GraphFramEx~\cite{GraphFramEx}, we define an explanation as an edge mask on the existing edges in the initial graph to be explained. We follow the original setting to train PGExplainer, GSAT, and RCExplainer. We further implement the diffusion-based explainer as introduced in Sec.~\ref{sec:generative_model}. D4Explainer generates an explanatory graph that can contain additional edges that are not in the initial graph. To keep consistent, we retrieve the common edges with the initial graph to evaluate D4Explainer in this work. GraphCFE is a simplified version of CLEAR~\cite{CLEAR} without the causality component, which is an explainability method for counterfactual explanations. Since we ignore the existence of any causal model in the datasets, we decide not to focus on the causality and use only the CLEAR-VAE backbone, \ie GraphCFE, in this work. We retrieve the important edges by subtracting the counterfactual explanation generated by GraphCFE from the initial graph. The remaining edges have weights of 1, while the rest have weights of 0.

\xhdr{Metrics} To evaluate the explainability methods, we use the systematic evaluation framework GraphFramEx~\cite{GraphFramEx}. We evaluate the methods on the faithfulness measure $fidelity-_{acc}$, which is defined as
$$fidelity-_{acc}= \frac{1}{N} \sum_{i=1}^{N}\left| \mathbbm{1}(\hat{Y}_f(G^i)=Y^i)- \mathbbm{1}(\hat{Y}_f(G^i_e)=Y^i) \right|,$$
where $G^i$ and $G^i_e$ denote the initial graph and the explanatory graph. $fidelity-_{acc}$ measures the proportion of the generated explanatory subgraphs that are faithful to the initial graph, leading to the same GNN prediction. 

\xhdr{Implementation} The experiments are run using the GraphFramEx code base available at \url{https://github.com/GraphFramEx/graphframex}. The code base evaluates generative and non-generative explainability methods on node and graph classification tasks over diverse synthetic and real-world datasets. 

% Following GraphFramEx~\cite{GraphFramEx}, we define an explanation as an edge mask on the existing edges in the initial graph to be explained. First, this constraint facilitates the comparison of very diverse explainability methods. Moreover, in the context of our study, all datasets are expected to be explained by some entities that already exist in the initial graphs, \ie motifs in synthetic datasets and groups of atoms in molecular datasets.

% It covers different explanation objectives in terms of explanation focus, size, and nature. Methods can be evaluated using the accuracy measure when the data has groundtruth explanations and faithfulness metrics, including fidelity+, fidelity- and the characterization score.

\subsection{Experimental Results}
\begin{figure}[h!]
    \centering
    \includegraphics[width=\textwidth]{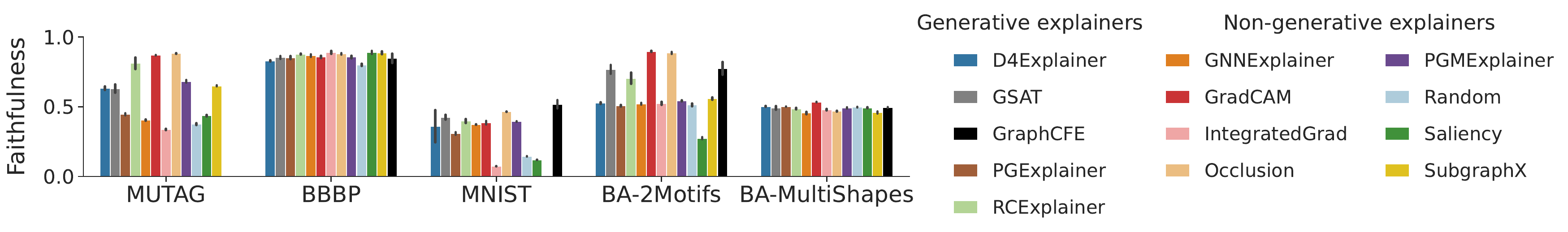}
    \caption{Faithfulness of explainability methods. Faithfulness score is computed as $1-fidelity-_{acc}$. On the x-axis, generative methods are always on the left-hand side of the methods (the bars). The faithfulness score is averaged over all explanations with less than 20 edges to enforce sparse and human-intelligible explanations.}
    \label{fig:faithfulness}
\end{figure}

\xhdr{Faithfulness} We conducted a comprehensive comparison of the faithfulness between generative and non-generative methods using three real-world datasets (BBBP, MUTAG, and MNIST) and two synthetic datasets (BA-2Motifs and BA-MultiShapes). The results, depicted in Figure~\ref{fig:faithfulness}, indicate that generative methods are generally performing the same or better than non-generative methods. Specifically, for MNIST, generative methods outperform non-generative methods across the board. In the cases of MUTAG and BA-2Motifs, the generative methods RCExplainer, GraphCFE, and GSAT closely follow Grad-CAM and Occlusion in terms of faithfulness. Regarding BBBP and BA-MultiShapes, both generative and non-generative methods exhibit similar results. Generally, generative methods achieve state-of-the-art performance on benchmark graph datasets. We then demonstrate that generative methods possess additional desirable properties, such as efficiency and generalization capacity, which make them more appealing than non-generative methods.
\vspace{-0.3cm}

\begin{figure}[h!]
    \centering
    \includegraphics[width=\textwidth]{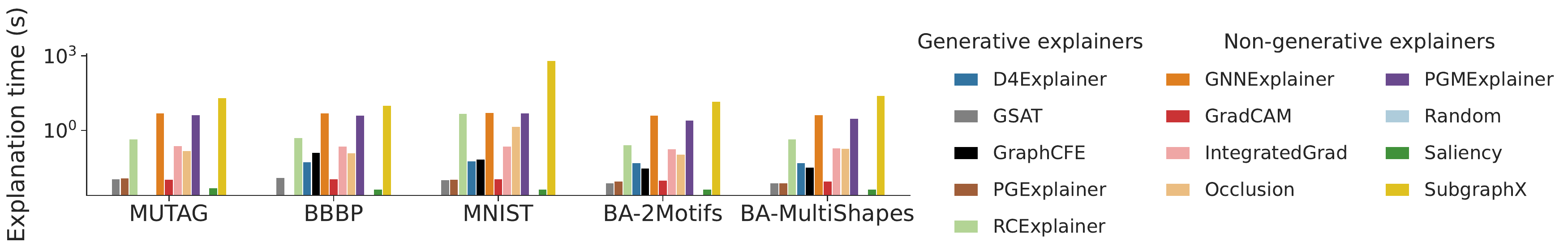}
    \caption{Inference time of explainability methods to explain one single graph. The computation time is averaged over 100 explanations over 5 seeds and reported in logarithmic scale.}
    \label{fig:efficiency}
\end{figure}

\xhdr{Efficiency} To measure the efficiency of explainability methods, we report the computation time to produce an explanation for a new instance in Figure~\ref{fig:efficiency}. Comparing generative methods with other learnable methods (\eg GNNExplainer, PGMExplainer) in Figure~\ref{fig:efficiency}, we observe that once the model is trained, generative explainability methods require shorter inference time than non-generative ones in general. The time is reported in logarithmic scale and generative methods always have inference times of the order of $10^0$ or less, except for the case of RCExplainer for MNIST. The advantage of shorter inference time is especially pronounced on large-scale datasets, \eg MNIST. We also report the time required to train a generative model from scratch in Table~\ref{tab:train_time}. 
% We report computation times both on CPUs (8 AMD EPYC 7742 CPUs with a memory of 5GB each) and on GPU (Nvidia GeForce RTX 2070 SUPER GPU with 8GB memory).
\begin{table}[h!]
\centering
\resizebox{0.8\linewidth}{!}{
\begin{tabular}{l|ccccc}
 & \textbf{D4Explainer} & \textbf{GraphCFE} & \textbf{GSAT} & \textbf{PGExplainer} & \textbf{RCExplainer} \\ \hline
\textbf{BA-2Motifs} & 475.3                  & 320.9             & 23.1          & 11.6                 & 194.0                \\ 
\textbf{BA-MultiShapes} & 309.3                  & 211.8             & 20.0          & 17.2                 & 251.0                \\ 
\textbf{BBBP}       & 385.6                  & 1350.0            & -             & 26.0                 & 303.4                \\ 
\textbf{MNIST}      & 934.6                  & 929.5             & 41.4          & 28.6                 & 3271.0               \\ 
\textbf{MUTAG}      & 253.1                  & -                 & 79.8          & 27.7                 & 434.6                \\ \hline
Mean                & 471.6                  & 703.1             & 41.1         & 22.2                 & 890.8               \\ 
\end{tabular}}
\caption{Training times (s) of the generative methods with 1 GPU (Nvidia GeForce RTX 2080)}
\label{tab:train_time}
\end{table}

\begin{figure}[h!]
    \centering
    \includegraphics[width=\textwidth]{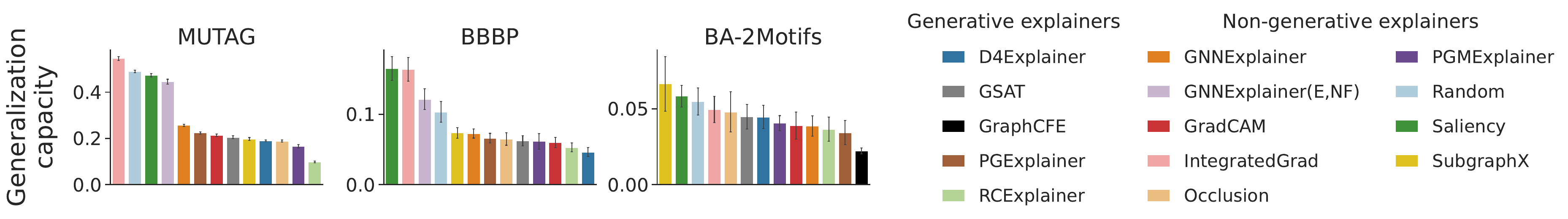}
    \caption{Generalization capacity of explainability methods is computed by subtracting the performance on data seen during training and the performance on unseen data. The lower the discrepancy reported on the y-axis, the better the method can generalize to unseen data. GNNExplainer indicates the explanations at the edge level and GNNExplainer(E,NF) represents the explanations for both edge existence and  node features.}
    \label{fig:generalization}\vspace{-0.5cm}
\end{figure}

\xhdr{Generalization} To compare generative and non-generative explainability methods on their generalization capacity, we split the datasets into seen and unseen data. The split ratio is 90/10$\%$. We further split the seen data into training, validation, and test set. The GNN model and the generative explainability methods are trained on the seen data. For non-generative methods, we explain 100 graphs from the seen dataset. Then, we test the trained methods on the unseen data. In Figure~\ref{fig:generalization}, we report the scores discrepancies between the test set of the seen data and the 10$\%$ unseen data for each explainability method. We also visualize the standard error on the five random seeds in Figure~\ref{fig:generalization}. Methods with higher absolute score discrepancies cannot generalize well to unseen data, while the ones with lower score discrepancies have a powerful generalization capacity. We can observe from Figure~\ref{fig:generalization} that generative explainability methods have lower scores than non-generative methods across three datasets in general, which demonstrates the better generalization capacity.

% \begin{figure}
%     \centering
%     \includegraphics[width=0.75\textwidth]{figures/top_balanced_acc_syn.pdf}
%     \caption{Caption}
%     \label{fig:top_acc_syn}
% \end{figure}

% \begin{figure}
%     \centering
%     \includegraphics[width=0.85\textwidth]{figures/fid+fid-acc.pdf}
%     \caption{Caption}
%     \label{fig:fid+/-}
% \end{figure}

% \begin{figure}
%     \centering
%     \includegraphics[width=0.85\textwidth]{figures/charact_acc.pdf}
%     \caption{Caption}
%     \label{fig:charact_topk}
% \end{figure}

% \subsection{Counterfactual explanations}

\section{Conclusion}
In this paper, we present a comprehensive review of explanation methods for Graph Neural Networks (GNNs) from the perspective of graph generation. By proposing a unified optimization objective for generative explanation methods, encompassing Attribution and Information constraints, we provide a framework to analyze and compare existing approaches. Moreover, we conduct a comparison between different approaches and empirically demonstrate the enhanced efficiency and generalizability of generative approaches compared to instance-dependent methods. Our study reveals shared characteristics and distinctions among current methods, laying the foundation for future advancements in the GNN explainability field.
% Overall, our work contributes to the advancement of transparent and trustworthy graph-based models, paving the way for improved outcomes in various applications through better feature extraction and understanding of complex graph-structured data.

\newpage

\bibliographystyle{plain}
\bibliography{reference.bib}


% \begin{thebibliography}{10}

% \end{thebibliography}

\end{document}